\documentclass[letterpaper]{article} 
\usepackage{aaai23}  
\usepackage{times}  
\usepackage{helvet}  
\usepackage{courier}  
\usepackage[hyphens]{url}  
\usepackage{graphicx} 
\usepackage{natbib}  
\usepackage{caption} 
\usepackage{algorithm}
\usepackage{algorithmic}

\usepackage{newfloat}
\usepackage{listings}
\DeclareCaptionStyle{ruled}{labelfont=normalfont,labelsep=colon,strut=off} 
\floatstyle{ruled}
\newfloat{listing}{tb}{lst}{}
\floatname{listing}{Listing}
\usepackage{booktabs}

\graphicspath{{figs}}

\title{PCTreeS --- 3D Point Cloud Tree Species Classification \\Using Airborne LiDAR Images}
\author{
    Hongjin Lin\equalcontrib,
    Matthew Nazari\equalcontrib,
    Derek Zheng\equalcontrib
}
\affiliations{
    John A. Paulson School of Engineering and Applied Sciences, Harvard University\\

    150 Western Ave, Allston, MA 02134\\
    hongjin\_lin@g.harvard.edu, \{matthewnazari, derek\_zheng\}@college.harvard.edu
}

\usepackage{bibentry}

\begin{document}

\maketitle

\begin{abstract}

Reliable large-scale data on the state of forests is crucial for monitoring ecosystem health, carbon stock, and the impact of climate change. Current knowledge of tree species distribution relies heavily on manual data collection in the field, which often takes years to complete, resulting in limited datasets that cover only a small subset of the world’s forests. Recent works show that state-of-the-art deep learning models using Light Detection and Ranging (LiDAR) images enable accurate and scalable classification of tree species in various ecosystems. While LiDAR images contain rich 3-Dimensional (3D) information, most previous works flatten the 3D images into 2D projections in order to use Convolutional Neural Networks (CNNs). This paper offers three significant contributions: 1) we apply the deep learning framework for tree classification in tropical savannas; 2) we use Airborne LiDAR images, which have a lower resolution but greater scalability than Terrestrial LiDAR images used in most previous works; 3) we introduce the approach of directly feeding 3D point cloud images into a vision transformer model (PCTreeS). Our results show that the PCTreeS approach outperforms current CNN baselines with 2D projections in AUC (0.81), overall accuracy (0.72), and training time ($\sim$ 45 mins). This paper also motivates further LiDAR image collection and validation for accurate large-scale automatic classification of tree species.
\end{abstract}

\section{Introduction}

Anthropogenic climate change, deforestation, and other human activities are well-known to impact ecological systems. To accurately measure these effects and design interventions, there is a need to gain a more accurate understanding of the state of forests. Currently, tree species censuses are collected by field experts manually, which could take years to finish. 

Airborne LiDAR (Light Detection and Ranging) images collected by Unmanned Aerial Vehicles (UAVs) are a fairly untapped source of data that provides a reliable, scalable, and cost-effective method for researchers to map tree species around the world. Recent development in computer vision enables the use of LiDAR images for automatic tree species classification. Previous works show that state-of-the-art computer vision models, especially Convolutional Neural Networks (CNNs), are highly accurate in classifying tree species using LiDAR images in various ecosystems \cite{zou2017tree, xi2020see, terryn2020tree, seidel2021predicting, allen, BUDEI2018632, HAMRAZ2019219, MAYRA2021112322, Hell2022}. 

This paper aims to develop an automatic classifier for tree species using LiDAR data collected at the Mpala Research Center, located on the Laikipia Plateau, Kenya, spanning over 48,000 acres of arid and semi-arid savannas and woodlands. The Mpala area is home to many of Africa’s distinctive large mammals, including elephants, lions, giraffes, and buffalos. The vegetation is dominated by legumes, particularly the genus \textit{Acacia}. A more accurate approximation of the forests at Mpala will have a profound impact on the scientific understanding of the ecosystem and its biodiversity.

Building on existing literature, we examine two deep-learning classification approaches. The first approach follows the existing work using 2D projections of LiDAR images with typical CNN models. The second approach applies a novel point cloud transformer (PCT) developed by
\citealt{guo2021} to classify 3D LiDAR images directly. We call the second approach PCTreeS (Point Cloud Transformer for Tree Species Classification). 

In this paper, we show that PCTreeS outperforms the baseline 2D CNN approach in AUC, overall classification accuracy, and training time. The transformer framework has a high potential for automatic tree species mapping with Airborne LiDAR images at scale.

\section{Previous Works and Contributions}

The availability of high-resolution LiDAR technology and recent development in computer vision enables unprecedented advancements in automatic tree species classification. A large body of recent works shows that deep learning models like CNNs, Random Forests (RFs), and Support Vector Machines (SVMs) achieve high classification accuracy \cite{micha2021, allen, seidel2021predicting, terryn2020tree, xi2020see}. This paper contributes to the current literature in three significant ways pertinent to ecosystems, the type of LiDAR images, and classification models (Table \ref{table:prevwork}). 

\subsection{Ecosystems}

Recent works focus primarily on woodland and forest ecosystems in Europe, North America, and China. We extend the deep learning approach to tropical savannas in Africa. Savannas are characterized by widely spaced trees, making the LiDAR images easier to collect and segment. Trees that can survive irregular rainfalls and long periods of drought, particularly of the genus \textit{Acacia}, thrive in tropical savannas. To our best knowledge, this paper is the first to classify tree species in the savanna landscape in Africa. 

\begin{table*}[t]
\centering
\begin{tabular}{cclrcc}\toprule
     & \multicolumn{1}{c}{LiDAR} & \multicolumn{1}{c}{Biome} & \multicolumn{1}{c}{Samples} & \multicolumn{1}{c}{Classes} & \multicolumn{1}{c}{Method} \\ \midrule
    \citealt{zou2017tree} & Terrestrial & Chinese Plantation Forest & $\sim$40,000 & 8 & 2D Deep Learning\\
    \citealt{xi2020see} & Terrestrial & Canadian, Finnish Woodland & 771 & 9 & 2D Deep Learning\\
    \citealt{terryn2020tree} & Terrestrial & UK Woodland & 758 & 5 & Support Vector Machines\\
    \citealt{seidel2021predicting} & Terrestrial & German, US Woodland & 690 & 8 & 2D Deep Learning\\
    \citealt{allen} & Terrestrial & Spanish woodland & 2,478 & 5 & 2D Deep Learning\\
    \citealt{BUDEI2018632} & Airborne & Canadian Plantation Forests & 1,658 & 10 & Random Forests \\
    \citealt{HAMRAZ2019219} & Airborne & US Robinson Forest & 3,987 & 2 & 2D Deep Learning \\
    \citealt{MAYRA2021112322} & Airborne & Finnish Southern Boreal Forests & 2,826 & 4 & 2D and 3D Deep Learning \\
    \citealt{Hell2022} & Airborne & Bavarian Forest National Park & 2,721 & 4 & 2D and 3D Deep Learning \\
    This paper & Airborne & Kenyan Tropical Savanna & $\sim$4,000 & 6 & 3D Vision Transformer \\ \bottomrule
\end{tabular}
\caption{Recent works on tree species classification with LiDAR images and deep learning models.}
\label{table:prevwork}
\end{table*}

\subsection{LiDAR Scanning Methods}

LiDAR images are 3D point clouds consisting of individual points that collectively form the shape of an object like a tree. There are two main types of LiDAR images — Terrestrial and Airborne. Terrestrial LiDAR images are collected by a scanner mounted on a stationary mechanism (e.g., a tripod) or slow-moving car to provide detailed scans of the surrounding areas. Terrestrial LiDAR scanning produces high-resolution point cloud images but is limited in scale. Airborne LiDAR techniques involve mounting a scanner on a flying drone to produce a larger scan of an area but produce relatively sparse point clouds. Airborne LiDAR works specifically well for ecosystems with sparser forests like the savannas. 

Recent works on tree species classification with deep learning techniques mainly rely on terrestrial LiDAR images. \citealt{zou2017tree}, \citealt{xi2020see}, \citealt{seidel2021predicting}, and \citealt{allen} are some recent works that train CNN models with 2D projections of 3D terrestrial LiDAR images and achieve high overall and species-wise classification accuracy. However, it is hard to scale the mapping approach to a large area due to the scale limitation of Terrestrial LiDAR data. Several works tabbed into Airborne LiDAR images and showed promising results using Random Forests, CNNs, and PointNet \cite{HAMRAZ2019219, BUDEI2018632, MAYRA2021112322, Hell2022}. Our paper contributes to the latter body of work and is the first to apply a novel point cloud vision transformer for the 3D Airborne LiDAR tree classification task. 

\subsection{Classification Models}

The most common deep learning approach to date is to feed 2D projections of 3D point clouds into an out-of-box CNN model. For example, in \citealt{allen}, six orthogonal projections are taken (2 vertical, 4 horizontal perspectives) to capture spatial data from 3D point clouds. However, these methods are inherently limited by the number of projections taken, as spatial information is lost when flattening 3D data into two dimensions. 

Several works explored the classification of 3D point clouds through deep learning approaches \cite{guo2019}, including PointNet++ \cite{pointnetplusplus}, which builds on PointNet \cite{pointnet}. PointNet was one of the first deep learning methods to approach classification by maintaining 3D point cloud structures, as opposed to 2D projections or other methods that manipulate the data into a separate form. PointNet++ offers a robust local feature extraction strategy that utilizes point neighborhood information at multiple scales.

Recent developments in 3D computer vision enable a whole host of techniques for LiDAR point cloud classification. Primarily, the significant development in 3D classification methods has centered around the Transformer, a technique that grew out of natural language processing and has since been adapted for computer vision tasks \cite{lu2022survey}. Transformers leverage a unique self-attention mechanism that enables efficient global input feature learning, and, subsequently, improves long-range dependency modeling as compared to CNNs. \citealt{guo2021} introduce the Point Cloud Transformer (PCT) for 3D point cloud classification tasks. PCT proposes a neighbor embedding module to encode spatial data from 3D point clouds into input embedding modules, improving global and local point cloud representation for classification. The authors also contribute an optimized Offset-Attention module which improves upon previous self-attention module implementations.

In this paper, we apply the PCT model developed by \citealt{guo2021} to classify 3D LiDAR images of trees directly. To our knowledge, this paper is the first to use a vision transformer framework for the 3D tree species classification task. 

\section{Data}
Our dataset $\mathcal{D} = \{(x_i, y_i)\}$ draws from two main sources. The ground-truth labels $\{y_i\}$ were provided by ForestGEO's Mpala plot census, and the Airborne LiDAR images $\{x_i\}$ were provided by the Davies Lab of Harvard University.

\subsection{Groud-truth Labels}
The Mpala plot census data was collected by ForestGEO from 2010 to 2015. In total, there are 136,752 trees (main stems) with 67 species labels spanning an area of 120 hectares. The census contains detailed information about individual stem locations, species, diameters at breast height (DBH), and status (alive or dead). 

\subsection{LiDAR Images} 
\begin{figure}
    \centering
    \includegraphics[width=0.7\columnwidth]{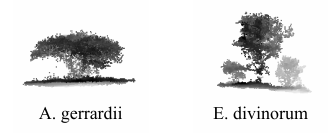}
    \caption{Examples of tree species that are observably discernable assuming very little noise.}
    \label{fig:species}
\end{figure}

We obtained a rich unlabelled dataset of Airborne LiDAR images collected by the Davies Lab in February 2022. These images are derived from a single composite scan of the Mpala plot within the ForestGEO census grid from a low-altitude (50m) UAV. The scan was further segmented into 43,709 individual trees as point cloud images of LiDAR by the Davies Lab, each with location and height information (Figure \ref{fig:species}).

\begin{figure}
    \centering
    \includegraphics[width=\columnwidth]{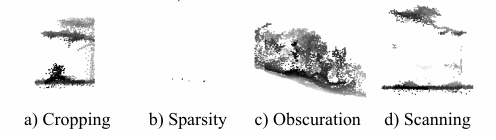}
    \caption{Noise in the dataset was caused by four main classes of error.}
    \label{bad:species}
\end{figure}

\subsection{Data Matching and Processing}

We matched ground-truth labels to LiDAR images using derived location information. Due to differences in the georeferencing systems in both datasets, only a subset of images was matched. The ForestGEO dataset decodes tree locations as their distance (due east and due north) from a corner post. We rely on domain experts from the Davies Lab to approximate the locations of the corner posts and boundaries of the census plot. We then use the Universal Transverse Mercator (UTM) coordinate system to decode the locations of each tree in both datasets. 

Since the geocoordinates of each label were recorded by hand in the field with respect to permanent grid stakes (which were themselves laid out by hand), there are inevitable data errors. Therefore, we allow some buffer in the matching process to account for human error and natural noise. The determination of the buffer depends on a tradeoff of match rate and accuracy. A large buffer enables more matches but may lead to inaccurate matchings. With valuable validation work from the Davies Lab, we round the UTM coordinates to the nearest ones to balance matching coverage and accuracy. This buffering technique, though naive, allows us to match about $\sim$4,000 LiDAR images with ground-truth labels for our classification models.

The resulting dataset contains 41 species of trees in total. The five most common species (\textit{Acacia drepanolobium}, \textit{Croton dichogamous}, \textit{Euclea divinorum}, \textit{Acacia brevispica}, and \textit{Acacia mellifera}) account for about 90.9\% of the matched images. To address the class imbalance issue, we group the remaining tree species into an ``other'' class for training, resulting in a total of 6 classes. Another way to balance the dataset would be to upsample the classes with fewer images and downsample the classes with more images.

\section{Methods}

\subsection{2D Projection-based Classification}

We follow \citealt{allen} to construct a baseline model using CNN. It leverages simultaneous multi-view perspective projections, while many previous works use a single 2D projection to capture each point cloud or use multiple projections but treat them as separate data points. The only approach that outperforms this method is \citealt{zou2017tree}. However, \citealt{zou2017tree} uses high-fidelity samples that are hard to achieve by subsequent data collection efforts.

\begin{figure}
    \includegraphics[width=\columnwidth]{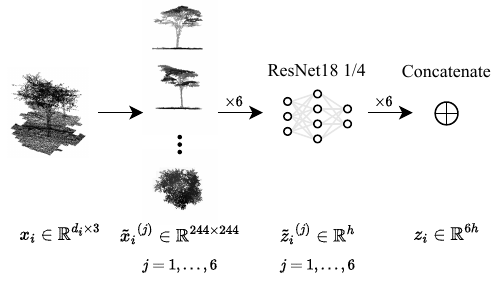}
    \caption{A diagrammatic representation of the baseline approach using 2D CNN following \citealt{allen}.}
    \label{fig:baseline}
\end{figure}

The baseline method preprocesses each point cloud by taking six orthogonal projections as inputs to a backbone CNN model. These 6 single-channel images are treated as separate data points, expanding the batch size by 6-fold. Before the final fully connected layer, the features produced by the convolutional layers are concatenated before being fed into a dense layer (Figure \ref{fig:baseline}).

The backbone CNN maintains a ResNet18 architecture with $1/4$ the filters and single channel input. Since this is a custom architecture, there were no pre-trained weights.

We hypothesize several improvements for this baseline approach. For example, to include information on the height of trees, we rescale each point cloud image by the same factor to maintain scale across all examples. In addition, we feed all six projections as channels of the same data point into the same ResNet18 architecture, since representations salient to classify tree species from a top-down perspective might not be the same as from a side perspective. Explicitly encoding that these projections are of the same data point could potentially help the model leverage the 3D information. We call this approach baseline++.

\subsection{3D Point Cloud Transformer}



The use of vision transformers has been shown to surpass traditional methods in 2D and 3D classification tasks \cite{lu2022survey}. \citealt{guo2021} introduces Point Cloud Transformer (PCT), a state-of-the-art framework for point cloud learning tasks, including image classification. PCT leverages a coordinate-based point embedding which converts 3D spatial data into a higher dimensional embedding space that maintains relative point similarity (i.e., the distance between points). While point embedding can effectively extract global features, a dedicated neighbor embedding is implemented to extract local features, leveraging past works such as PointNet++ \cite{pointnetplusplus}. PCT also includes an Offset-Attention module that improves upon the original transformer self-attention module. The model was evaluated on ModelNet40, a dataset commonly used for point cloud shape classification tasks, and was shown to outperform other state-of-the-art methods. 

This paper implements PCT on our dataset by feeding pure point cloud images of labeled LiDAR images into the model. Notably, there are no projections used, and the original 3D spatial data is maintained in the input.

\section{Experiments and Results}

We trained three main models: baseline, baseline++, and PCTreeS. As mentioned in the Methods section, the baseline model is a ResNet 18 1/4 network with 2D projections of each LiDAR image. The baseline++ model builds upon the baseline model and allows height normalization and single-channel inputs, which we hypothesize to be beneficial for training. The PCTreeS model takes in 3D point clouds and performs the classification tasks without further processing.

To mitigate concerns with class imbalances, we set the models to learn 6 classes of tree species which comprised the top 5 most common species, and an ``other'' class to capture the remaining tree species. In addition, we trained models on images with over 1,000 points as a heuristic for filtering out sparse LiDAR scans that we hypothesized would yield poor accuracy.  

All three models are trained on the same technical setup on one GPU (Dell DSS 8440 Cauldron). We keep model parameters largely the same across all three models, e.g., batch size 32, epoch 100, learning rate $1e-5$, and random seed. 

All three models achieve decent performance within a short training period, which is important for computing sustainability and reproducibility. PCTreeS outperforms the baseline model with CNN in all three performance metrics (AUC, overall accuracy, and training time) (Table \ref{table:performance}). While our best performance in overall accuracy is lower than \citealt{allen} (81\%), we believe the difference is mainly driven by differences in point density between Terrestrial and Airborne LiDAR images and various data issues in the tree segmentation process. The 3D point cloud transformer framework is a promising step towards more accurate mapping of tree species at scale.

\begin{table}[t]
\centering
\begin{tabular}{cclrcc}\toprule
      & \multicolumn{1}{c}{AUC} & \multicolumn{1}{c}{Accuracy} & \multicolumn{1}{c}{Training Time} \\ \midrule
    Baseline & 0.75 & 0.68 & $\sim$90 mins \\
    Baseline++ & 0.75 & 0.70 & $\sim$90 mins \\
    PCTreeS & 0.81 & 0.72 & $\sim$45 mins\\
    \bottomrule
\end{tabular}
\caption{Model performance. AUC and overall accuracy are taken from the last epoch.}
\label{table:performance}
\end{table}

\section{Conclusion}

This paper examines two approaches to tree species classification in tropical savannas. The first method uses a CNN trained on 2D projections of LiDAR images, and the second approach leverages a state-of-the-art 3D point cloud vision transformer (PCTreeS). We show that PCTreeS outperforms state-of-the-art CNN models with 2D projections while reducing training time significantly.

We see many ways to improve upon the current model performance. First, we take the segmented individual LiDAR images as a given, but the quality of the segmented tree images can be improved as a separate task. Currently, we see a notable portion of images with either too few points to form a tree (in single digits of points), multiple trees, cropping errors, and non-tree objects like bushes (Figure \ref{bad:species}). Proper isolation of individual trees is key to achieving accurate classification results. We are working with the Davies lab to examine these data issues further and improve the segmentation accuracy. Second, data augmentation has been proven to be valuable in improving classification accuracy for terrestrial LiDAR images \cite{allen}. We plan to incorporate more intensive data augmentation and processing to generate a richer dataset for the training step.

Furthermore, situated in the emerging field of AI for Social Impact, our research establishes that close collaborations with domain experts are crucial for developing accurate and helpful AI-powered applications. Without the support of the Davies lab, we would not have access to the LiDAR data nor the domain insights on how to interpret the images. The ecological context was especially helpful for us to incorporate information useful for tree species classification, such as tree height. 

Finally, to facilitate future replication of this paper and future work, we have made the repository public.\footnote{Available at \url{https://github.com/mattynaz/pctrees}} For data access, please contact the authors and Davies lab for the LiDAR images and directly request the Mpala plot census data using the ForestGEO data request form available online. 

\section{Acknowledgements}

We thank the Davies Lab, especially Tyler Coverdale and Peter Boucher, for their invaluable expertise and support throughout the project. We would also like to thank Sonja Johnson-Yu and Paula Rodriguez Diaz for their helpful questions and comments helped us improve the quality of this project.

We would also like to acknowledge the incredible data collection work by ForestGEO and their collaborators: the 120-ha Mpala plot is a collaborative project of the National Museums of Kenya, the Kenya Wildlife Service, and the Mpala Wildlife Foundation, in partnership with the Forest Global Earth Observatory (ForestGEO). Funding for the first census was provided by the Center for Tropical Forest Science--Forest Global Earth Observatory.

\bibliography{pctrees}

\end{document}